\documentclass[%
superscriptaddress,
preprint,
 amsmath,amssymb,
 aps,
]{revtex4-1}

\usepackage{graphicx}
\usepackage{dcolumn}
\usepackage{bm}
\usepackage{xcolor}
\usepackage{qcircuit}
\usepackage{braket}
\usepackage{tikz}
\usepackage[flushleft]{threeparttable}
\usepackage{hyperref}

\renewcommand{\figureautorefname}{Figure~\negthinspace}
\renewcommand{\equationautorefname}{Equation~\negthinspace}
\renewcommand{\tableautorefname}{Table~\negthinspace}
\renewcommand{\sectionautorefname}{Section~\negthinspace}

\begin{document}

\preprint{BNL-SBU}

\title{Hybrid Quantum-Classical Graph Convolutional Network}

\author{Samuel Yen-Chi Chen}
\email{ychen@bnl.gov}
\affiliation{Computational Science Initiative, Brookhaven National Laboratory, Upton, NY 11973, USA}%

\author{Tzu-Chieh Wei}
\email{tzu-chieh.wei@stonybrook.edu}
\affiliation{C. N. Yang Institute for Theoretical Physics and Department of Physics and Astronomy, State University of New York at Stony Brook, Stony Brook, NY 11794-3840, USA}

\author{Chao Zhang}
\email{czhang@bnl.gov}
\affiliation{Physics Department, Brookhaven National Laboratory, Upton, NY 11973, USA}

\author{Haiwang Yu}
\email{hyu@bnl.gov}
\affiliation{Physics Department, Brookhaven National Laboratory, Upton, NY 11973, USA}

\author{Shinjae Yoo}%
 \email{sjyoo@bnl.gov}
\affiliation{Computational Science Initiative, Brookhaven National Laboratory, Upton, NY 11973, USA}

\date{\today}

\begin{abstract}
 
The high energy physics (HEP) community has a long history of dealing with large-scale datasets. To manage such voluminous data, classical machine learning and deep learning techniques have been employed to accelerate physics discovery. Recent advances in quantum machine learning (QML) have indicated the potential of applying these techniques in HEP. 
However, there are only limited results in QML applications currently available. In particular, the challenge of processing sparse data, common in HEP datasets, has not been extensively studied in QML models.
This research provides a hybrid quantum-classical graph convolutional network (QGCNN) for learning HEP data. 
The proposed framework demonstrates an advantage over classical multilayer perceptron and convolutional neural networks in the aspect of number of parameters. Moreover, in terms of testing accuracy, the QGCNN shows comparable performance to a quantum convolutional neural network on the same HEP dataset while requiring less than $50\%$ of the parameters.
Based on numerical simulation results, studying the application of graph convolutional operations and other QML models may prove promising in advancing HEP research and other scientific fields.
\end{abstract}

\maketitle


\section{\label{sec:Indroduction}Introduction}
%
%
The high energy physics (HEP) community has a long tradition of processing large-scale datasets. Recent advances in machine learning (ML) and deep learning (DL) techniques have introduced many new valuable concepts and tools to augment HEP research \cite{baldi2014searching, guest2018deep, baldi2016jet, baldi2016parameterized, guest2016jet, de2016jet}. For example, convolutional neural networks (CNN) have been transformative in streamlining analysis of large HEP datasets \cite{aurisano2016convolutional, abi2020neutrino}. Meanwhile, recent progress in graph convolutional neural networks (GCN) has helped manage the difficulties in processing sparse data \cite{wu2020comprehensive, zhou2018graph, zhang2018graph}, which is ubiquitous in HEP.  

In parallel with the advancements in ML/DL, quantum computers, once cited as ``impractical,'' have been built by several companies \cite{arute2019quantum, cross2018ibm, grzesiak2020efficient}. In theory, quantum computing can solve certain problems that are unworkable using classical computers \cite{harrow2017quantum, nielsen2002quantum, shor1999polynomial, grover1997quantum}. 
%
However, currently available quantum devices, the so-called \textit{noisy intermediate-scale quantum} (NISQ) processors~\cite{preskill2018quantum}, are not capable of performing robust quantum computing with many numbers of qubits and large circuit depth due to the lack of quantum error correction. Thus, it is non-trivial to design a proper hybrid quantum-classical architecture that can harness the strength and scalability of both computing paradigms. For clarity, the term ``hybrid'' in this case represents using classical computers for optimization and quantum computers for certain complicated tasks.
%

Despite limits on the number of available qubits and circuit depth, numerous efforts have sought to design ML applications on NISQ devices. Indeed, a family of algorithms called \emph{variational quantum algorithms} \cite{cerezo2020variational}, which have been successful in calculating chemical ground states \cite{cerezo2020variational, peruzzo2014variational}, have achieved promising results in quantum machine learning (QML) \cite{schuld2018circuit, Farhi2018ClassificationProcessors, benedetti2019parameterized, mari2019transfer, abohashima2020classification, easom2020towards, sarma2019quantum, chen2020hybrid, stein2020hybrid,chen2020quantum,chen2020qcnn,kyriienko2020solving,dallaire2018quantum, stein2020qugan, zoufal2019quantum, situ2018quantum,nakaji2020quantum,lloyd2020quantum, nghiem2020unified,chen19, lockwood2020reinforcement,wu2020quantum, jerbi2019quantum, Chih-ChiehCHEN2020,bausch2020recurrent,yang2020decentralizing}.
%
%
Yet, certain problems have not been thoroughly studied under current QML techniques. For example, sparse data, which is common in scientific data, especially within the HEP community, generally is difficult for ML models, and it is unclear if current QML models can provide advantages in addressing this problem. 
%
In classical ML, one potential solution for dealing with sparse data is by incorporating graph convolutional operations in DL models. However, this has not been thoroughly investigated in the quantum domain.

This work presents a novel hybrid quantum-classical graph convolutional neural network (QGCNN) framework to demonstrate the quantum advantage over classical algorithms.
Contributions stemming from this work include:
\begin{itemize}
    \item Successfully demonstrate the hybrid model with graph convolutional operation and variational quantum circuits.
    \item Illustrate the superior performance in terms of testing accuracy over the classical multilayer perceptron (MLP) model and classical convolutional neural networks (CNN).
    \item Showcase the comparable performance in terms of testing accuracy to quantum convolutional neural networks (QCNN) on the same Deep Underground Neutrino Experiment (DUNE) dataset while requiring less than $50\%$ of the model parameters.
\end{itemize}
 In this paper, Section~\ref{sec:HighEnergyPhysicsData} introduces the HEP experimental data used in this work. In Section~\ref{sec:GraphConvolution}, \ref{sec:VariationalQuantumCircuit} and \ref{sec:GraphConvAndVQC} describe the new QGCNN architecture in detail. Section~\ref{sec:ExpAndResults} shows the QGCNN's performance on the experimental data, followed by additional discussions in Section~\ref{sec:Discussion}. Finally, Section~\ref{sec:Conclusion} includes the concluding details.

\section{\label{sec:HighEnergyPhysicsData}Training and testing dataset}
This work uses the same simulated data as our team's previous work in employing QCNN for HEP event classification~\cite{chen2020qcnn}. The dataset is simulated for the DUNE experiment~\cite{dune} with the Wire-Cell Toolkit~\cite{wct} and LArSoft software~\cite{larsoft}. By using the same dataset, we can compare our previous results to benchmark the performance of the new QGCNN algorithms. While details of the experiment and data simulation can be found in Ref.~\cite{chen2020qcnn}, we provide a brief description for completion (as follows).

\begin{figure}[htbp]
\centering
\includegraphics[width=0.85\linewidth]{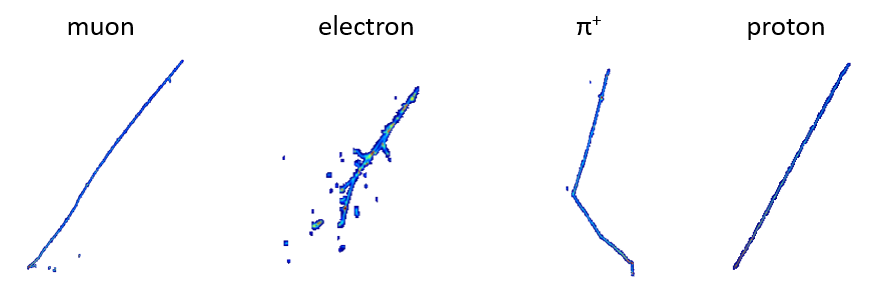}
\caption{Example images of simulated particle activities ($\mu^+$, $e^-$, $\pi^+$, $p$) in a LArTPC detector. Colors in the images represent the intensity of the ionization energy loss recorded by each pixel.}
\label{simu_particle}
\end{figure}

The DUNE experiment is a long-baseline neutrino oscillation experiment to search for CP violation in the lepton sector, determine neutrino mass ordering, perform precision tests of the three-neutrino paradigm, detect supernova neutrino bursts, and search for nucleon decays beyond the Standard Model. The experiment currently is under construction and will start taking data in the next few years.
The DUNE detector uses the Liquid Argon Time Projection Chamber (LArTPC) technology, which digitally records high-resolution images of particle activities~\cite{rubbia77,Chen:1976pp,willis74,Nygren:1976fe} in the detector. The training and testing dataset used in this work is generated with a full detector simulation of DUNE~\cite{wct,larsoft}. Single-particle images are generated by applying a realistic simulation of particle interaction, detector response, and digital signal processing~\cite{lartpc-sp}. 
Four different types of particles ($\mu^+$, $e^-$, $\pi^+$, and $p$) are simulated. \figureautorefname{\ref{simu_particle}} shows example images of simulated particles. The images have a resolution of $480\times600$ pixels, where each pixel represents approximately $5\times5$ square millimeter spacially. Each particle's momentum is set such that the mean range of the particle is about 2 meters, so the classification is not sensitive to the image size. Because of differences in the mass, charge, and interaction types of the particles, the particles leave rather distinctive topological patterns in the recorded images as shown in~\figureautorefname{\ref{simu_particle}}. Details about the underlining physics can be found in Ref.~\cite{chen2020qcnn}.  Similar to our QCNN work~\cite{chen2020qcnn}, the QGCNN algorithm seeks to classify the types of these different particles.

\section{\label{sec:GraphConvolution}Graph Convolution}
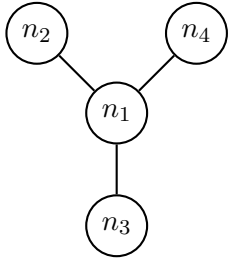
\begin{figure}[htbp]
\begin{tikzpicture}[node distance={15mm}, thick, main/.style = {draw, circle}] 
	\node[main] (1) {$n_1$}; 
	\node[main] (2) [above left of=1] {$n_2$}; 
	\node[main] (3) [below  of=1] {$n_3$}; 
	\node[main] (4) [above right of=1] {$n_4$}; 
	\draw (1) -- (2); 
	\draw (1) -- (3); 
	\draw (1) -- (4); 
\end{tikzpicture}
\caption[Example Graph.]{{\bfseries Example Graph.}
}
\label{Fig:ExampleGraph}
\end{figure}

\begin{figure}[htbp]
\begin{tikzpicture}[node distance={15mm}, thick, main/.style = {draw, circle}] 
	\node[main] (1) {$n_1$}; 
	\node[main] (2) [above left of=1] {$n_2$}; 
	\node[main] (3) [below  of=1] {$n_3$}; 
	\node[main] (4) [above right of=1] {$n_4$}; 
	\draw (1) -- (2); 
	\draw (1) -- (3); 
	\draw (1) -- (4); 
	\draw (1) to [out=165,in=255,looseness=5] (1);
	\draw (2) to [out=180,in=90,looseness=5] (2);
	\draw (3) to [out=225,in=315,looseness=5] (3);
	\draw (4) to [out=0,in=90,looseness=5] (4);
\end{tikzpicture}
\caption[Example Graph.]{{\bfseries Example Graph with Self Loop.}
}
\label{Fig:ExampleGraphSelfLoop}
\end{figure}
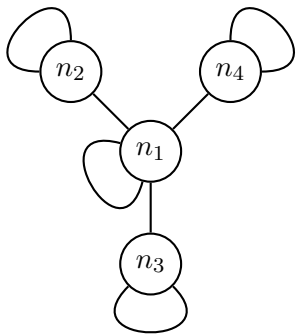

A graph is an ordered pair $G = (V,E)$, where $V$ is the set of nodes and $E$ is the set of edges. The \emph{adjacency matrix} $A$ of an undirected graph $\mathcal{G}$ with $N$ nodes $\{u_1 \cdots u_N\}$ is an $N \times N$ matrix with the property that the element $A_{ij} = 1$ if there is an edge between node $u_i$ and $u_j$ and is $0$ otherwise.
The \emph{normalized adjacency matrix} $\mathcal{A}$ is defined to be
\begin{equation}
\mathcal{A} = D^{-1 / 2} A D^{-1 / 2},
\end{equation}
where $D = diag(d)$ for $d(i)$, the degree of node $i$.
For an $N$-node graph $G$, the corresponding $D^{-1/2}$ is
\begin{equation}
D^{-1 / 2}=\left(\begin{array}{cccc}\frac{1}{\sqrt{d(1)}} & 0 & \cdots & 0 \\ 0 & \frac{1}{\sqrt{d(2)}} & \cdots & 0 \\ \vdots & \vdots & \ddots & \vdots \\ 0 & 0 & \cdots & \frac{1}{\sqrt{d(N)}}\end{array}\right).
\end{equation}
Consider the graph with four nodes $\{n_1, n_2, n_3, n_4\}$ shown in \figureautorefname{\ref{Fig:ExampleGraph}}, and, on each node, there is a corresponding feature value $f_i$ with $i = \{1,2,3,4\}$.
The feature vector $X$ for this graph is
\begin{equation}
    X = \left( \begin{matrix} f_1 \\ f_2 \\ f_3 \\ f_4 \end{matrix} \right).
\end{equation}
The adjacency matrix $A$ for this graph is
\begin{equation}
    A = \left( \begin{matrix} 0 & 1 & 1 & 1 \\ 1 & 0 & 0 & 0 \\ 1 & 0 & 0 & 0 \\ 1 & 0 & 0 & 0 \end{matrix} \right).
\end{equation}
The \emph{graph convolution} operation here is the matrix multiplication $AX$:
\begin{equation}
    AX = \left( \begin{matrix} 0 & 1 & 1 & 1 \\ 1 & 0 & 0 & 0 \\ 1 & 0 & 0 & 0 \\ 1 & 0 & 0 & 0 \end{matrix} \right)\left( \begin{matrix} f_1 \\ f_2 \\ f_3 \\ f_4 \end{matrix} \right) = \left( \begin{matrix} f_2+f_3+f_4 \\ f_1 \\ f_1 \\ f_1 \end{matrix} \right).
\end{equation}
In this example graph, the features of neighboring nodes aggregate together. When considering numerical computation, it is better to use the normalized adjacency matrix to avoid numerical instability (e.g., exploding values).
In addition, we may want to modify the $A$ to $\hat{A} = A + I$ in order to keep their individual features. This is equivalent to adding a \emph{loop} for each node (\figureautorefname{\ref{Fig:ExampleGraphSelfLoop}}),
Now, the adjacency matrix $A$ becomes $\hat{A}$, which is
\begin{equation}
    \hat{A} = A + I = \left( \begin{matrix} 1 & 1 & 1 & 1 \\ 1 & 1 & 0 & 0 \\ 1 & 0 & 1 & 0 \\ 1 & 0 & 0 & 1 \end{matrix} \right).
\end{equation}
Therefore, the aggregation operation is
\begin{equation}
    \hat{A}X = \left( \begin{matrix} 1 & 1 & 1 & 1 \\ 1 & 1 & 0 & 0 \\ 1 & 0 & 1 & 0 \\ 1 & 0 & 0 & 1 \end{matrix} \right)\left( \begin{matrix} f_1 \\ f_2 \\ f_3 \\ f_4 \end{matrix} \right) = \left( \begin{matrix} f_1 + f_2+f_3+f_4 \\ f_1 + f_2 \\ f_1 + f_3 \\ f_1 + f_4 \end{matrix} \right).
\end{equation}

Consider an image with the size of $N \times N$. It can be viewed as a graph with $N^2$ nodes. Such a graph is \emph{regular} because all nodes (pixels) of the graph are connected to each other in exactly the same manner. We define the adjacency matrix for an image based on the intuition that nearby nodes (or pixels) should have stronger relationships, while distant ones should have relatively weak relationships. For example, in a natural image, neighboring pixels are highly possible in the same object or architecture.
The adjacency matrix $A$ for this $N \times N$ image has the dimension $N^2 \times N^2$. Each element of $A$ is calculated according to:
\begin{equation}\label{eqn:adjacency_a_ij}
    A_{ij} = \exp{\left[\frac{-d_{ij}}{\sigma^{2}} \right]},
\end{equation}
where $d_{ij}$ represents the Euclidean distance for the node pair $(x_i, y_i)$ and $(x_j, y_j)$ with the value $d_{ij} = \sqrt{(x_i - x_j)^2 + (y_i - y_j)^2}$. The parameter $\sigma$ is the Gaussian scale. In this experiment, the value for $\sigma$ is $0.05 \times \pi$.

\begin{figure}[htbp]
\centering
\includegraphics[width=0.5\linewidth]{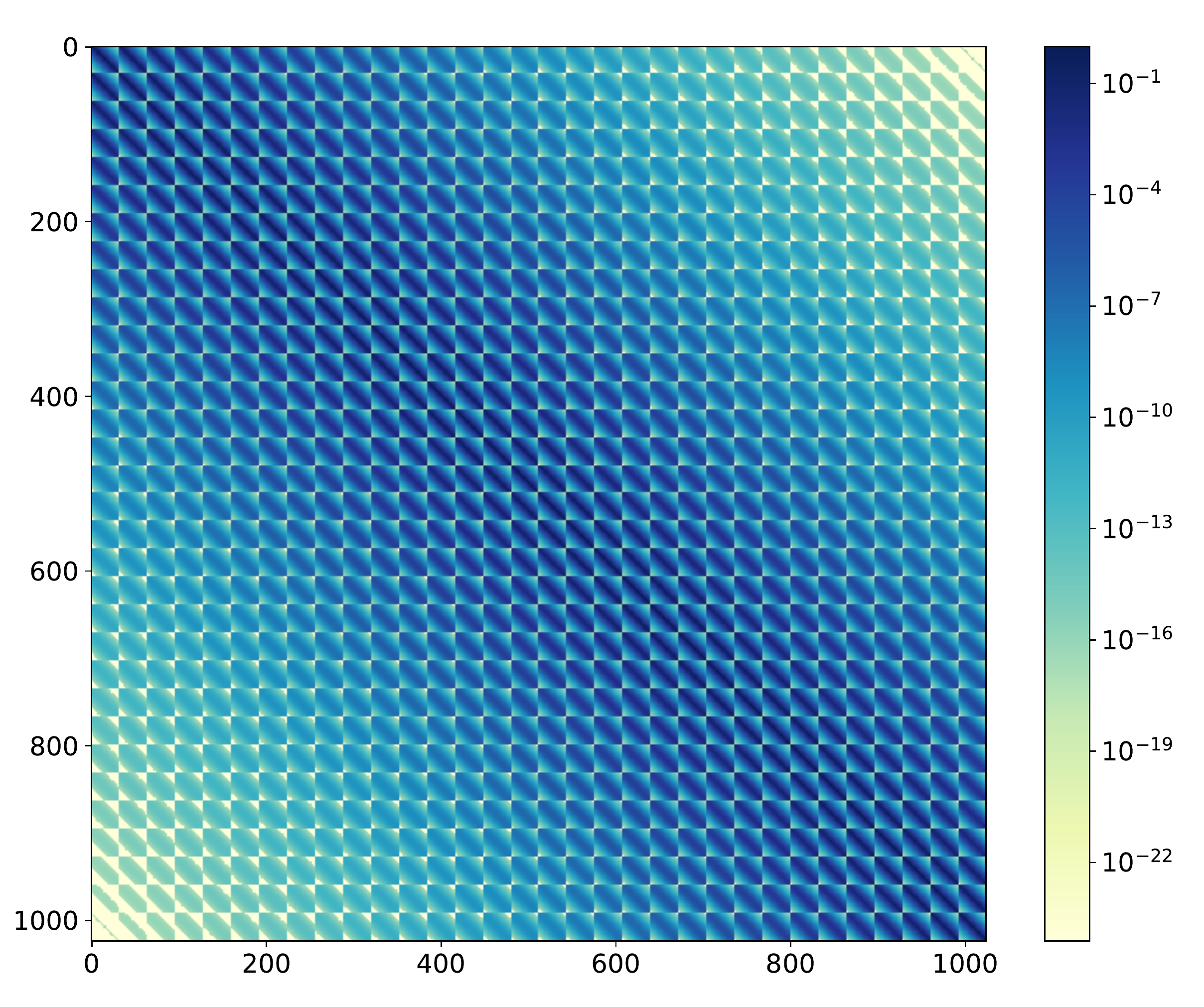}
\caption{{\bfseries Adjacency matrix $A$ for the $32 \times 32$ image.} }
\label{adjacency_matrix_1024_1024}
\end{figure}

\begin{figure}[htbp]
\centering
\includegraphics[width=1\linewidth]{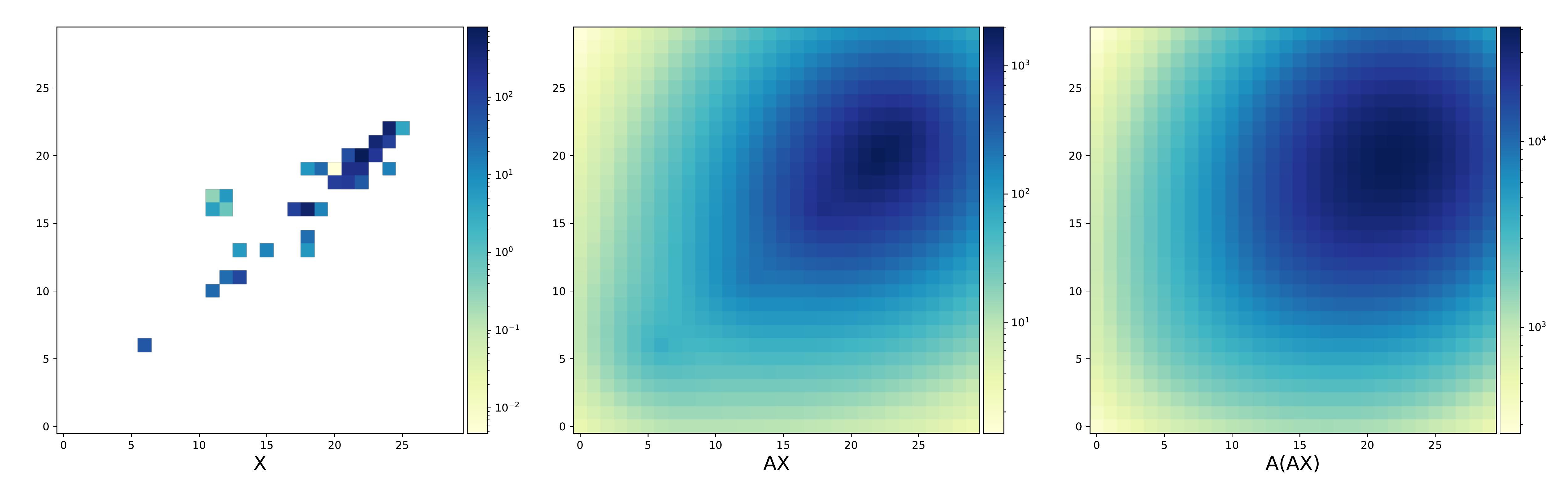}
\caption{{\bfseries Examples of graph convolution on the DUNE data.} An example image from the DUNE dataset used in this study. The original image $X$ is first flattened and multiplied by the adjacency matrix $A$. The transformed vector $AX$ then is reshaped to the original image format. This depicts the result of $AX$ and $A^2X$.}
\label{example_AX_AAX}
\end{figure}

For a $32 \times 32$ image, we can easily calculate the $1024 \times 1024$ dimension matrix $A$, and we can present this matrix as shown in \figureautorefname{\ref{adjacency_matrix_1024_1024}}.
We observe that the matrix values are much higher in the diagonal regions, corresponding to the fact that these points represent the node distances between nearby graph nodes.
Here, we consider an example from the training set. In \figureautorefname{\ref{example_AX_AAX}}, we demonstrate the effects of adjacency matrix $A$ (defined in \equationautorefname{\ref{eqn:adjacency_a_ij}}) on the input image $X$ from the DUNE-simulated dataset. The original $X$ is rather sparse, making it difficult for QML models to classify. The situation worsens when encoding the image with amplitude encoding (described in \sectionautorefname{\ref{sec:AmplitudeEncoding}}) as the vector normalization procedure causes significant information loss. 

\section{\label{sec:VariationalQuantumCircuit}Variational Quantum Circuits}
Variational quantum circuits (VQC) are a special kind of circuit with parameters that are adjustable via optimization procedures developed by the classical ML community. This family of algorithms was first developed to calculate chemical ground states \cite{peruzzo2014variational} and has been widely used~\cite{cerezo2020variational}. VQCs also are known as ``quantum neural networks,'' or QNN, when applied in the ML field.
Recent results have demonstrated that VQCs are more expressive than classical neural networks~\cite{sim2019expressibility,lanting2014entanglement,du2018expressive, abbas2020power} with respect to the number of parameters or learning speed.
Recent advances in VQC have demonstrated various applications in QML. For example, VQC has shown to be successful in the task of classification \cite{mitarai2018quantum, schuld2018circuit, Farhi2018ClassificationProcessors, benedetti2019parameterized, mari2019transfer, abohashima2020classification, easom2020towards, sarma2019quantum, stein2020hybrid, chen2020hybrid,chen2020qcnn,wu2020application}, function approximation \cite{chen2020quantum, mitarai2018quantum,kyriienko2020solving}, generative ML \cite{dallaire2018quantum, stein2020qugan, zoufal2019quantum, situ2018quantum,nakaji2020quantum}, metric learning \cite{lloyd2020quantum, nghiem2020unified}, deep reinforcement learning \cite{chen19, lockwood2020reinforcement, jerbi2019quantum, Chih-ChiehCHEN2020,wu2020quantum}, sequential learning \cite{chen2020quantum, bausch2020recurrent}, and speech recognition \cite{yang2020decentralizing}.
For a VQC-based model to process classical data, it must first encode the classical data into a quantum state.
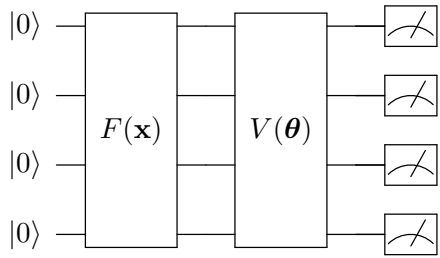
\begin{figure}[htbp]
\begin{center}
\begin{minipage}{10cm}
\Qcircuit @C=1em @R=1em {
\lstick{\ket{0}} & \multigate{3}{F(\mathbf{x})}  & \qw        & \multigate{3}{V(\boldsymbol{\theta})}       & \qw      & \meter \qw \\
\lstick{\ket{0}} & \ghost{F(\mathbf{x})}         & \qw        & \ghost{V(\boldsymbol{\theta})}              & \qw      & \meter \qw \\
\lstick{\ket{0}} & \ghost{F(\mathbf{x})}         & \qw        & \ghost{V(\boldsymbol{\theta})}              & \qw      & \meter \qw \\
\lstick{\ket{0}} & \ghost{F(\mathbf{x})}         & \qw        & \ghost{V(\boldsymbol{\theta})}              & \qw      & \meter \qw \\
}
\end{minipage}
\end{center}
\caption[General structure for the variational quantum circuit.]{{\bfseries General structure for the variational quantum circuit.}
The $F(\mathbf{x})$ is the quantum operation for encoding the classical data into the quantum state and $V(\boldsymbol{\theta})$ is the VQC block with the adjustable parameters $\boldsymbol{\theta}$. After the quantum operation, the quantum state is \emph{measured} to retrieve classical numbers for additional processing. The additional processing may be a classical neural network or another variational quantum circuit.
}
\label{Fig:GeneralVQC}
\end{figure}
A general $N$-qubit quantum state can be represented as:
\begin{equation}
\label{eqn:quantum_state_vec}
    \ket{\psi} = \sum_{(q_1,q_2,...,q_N) \in \{ 0,1\}^N}^{} c_{q_1,...,q_N}\ket{q_1} \otimes \ket{q_2} \otimes \ket{q_3} \otimes ... \otimes \ket{q_N},
\end{equation}
where $ c_{q_1,...,q_N} \in \mathbb{C}$ is the \emph{amplitude} of each quantum state and $q_i \in \{0,1\}$. 
The square of the amplitude $c_{q_1,...,q_N}$ represents the \emph{probability} of measurement with the post-measurement state in  $\ket{q_1} \otimes \ket{q_2} \otimes \ket{q_3} \otimes ... \otimes \ket{q_N}$, and the total probability should sum to unity, i.e.,
\begin{equation} 
\label{eqn:quantum_state_vec_normalization_condition}
\sum_{(q_1,q_2,...,q_N) \in \{ 0,1\}^N}^{} ||c_{q_1,...,q_N}||^2 = 1. 
\end{equation}
%
There are several different kinds of encoding methods that provide distinct quantum advantages with varying difficulties in the hardware implementation \cite{Schuld2018InformationEncoding, sierra2020tensorflow}. Several recent advances suggest that the encoding operation itself can be learned from the dataset \cite{lloyd2020quantum, nghiem2020unified}. The following sections introduce the two encoding schemes used in this work: \emph{amplitude encoding} and \emph{variational encoding}.
\subsection{\label{sec:AmplitudeEncoding}Amplitude Encoding}
In the first VQC block, we employ \emph{amplitude encoding} to reduce the number of qubits and circuit parameters. Here, we have a classical vector in the form of $(\alpha_{0} \cdots \alpha_{2^n-1})$. The aim is to encode it into an $n$-qubit quantum state $\ket{\Psi} = \alpha_{0}\ket{00\cdots 0} + \cdots + \alpha_{2^n-1}\ket{11\cdots 1}$, where the $\alpha_i \in \mathbb{R}$ and the vector $(\alpha_{0} \cdots \alpha_{2^n-1})$ is a normalized vector (summed to unity). The details about this encoding method are introduced in the work of~\cite{mottonen2005transformation} and also can be found in the textbook \cite{Schuld2018InformationEncoding}. The main reason to choose this encoding scheme is to minimize the number of qubits used, thereby reducing the number of circuit parameters. For example, given a vector with size $N$, it can be represented with a $\log_{2}(N)$-qubit system with amplitude encoding. In this work, we consider the input vector with dimension $N = 32 \times 32 = 1024$ and $n$-qubit system with $n = 10$.

\subsection{\label{sec:VariationalEncoding}Variational Encoding}
In the second VQC block, we employ \emph{variational encoding}, where the input values are used as the quantum rotation angles. 
In variational encoding, there is a predefined sequence of single-qubit rotation gates for each qubit. A single-qubit gate with rotation along the $j$-axis by angle $\theta$ is given by
\begin{equation}
  \label{eq:SingleQubitRotation}
  R_j(\theta)=e^{-i\theta\sigma_j/2}=\cos\frac{\theta}{2} I-i\sin\frac{\theta}{2}\sigma_j,
\end{equation}
where $I$ is the identity matrix and $\sigma_{j}$ is the Pauli matrix with $j = x, y, z$.
The \emph{rotation angles} $\theta$ are calculated from the input values. In this study, we choose $R_y$ and $R_z$ to encode the classical data with the rotation angles $\arctan(x)$ and $\arctan(x^2)$, respectively. 
Given an $n$-dimensional vector $\mathbf{x} = (x_{1},x_{2},\cdots,x_{n})$ to be encoded in an $n$-qubit circuit, the encoding operation can be written as
\begin{equation}
    U(\mathbf{x}) = R_{z}(\arctan(x_{1}^2))R_{y}(\arctan(x_{1})) \otimes \cdots \otimes R_{z}(\arctan(x_{n}^2))R_{y}(\arctan(x_{n})).
\end{equation}
The circuit of this encoding is presented in \figureautorefname{\ref{Fig:VQCInterBlock}}.
\subsection{Optimization of Quantum Circuits}
For gradient-based optimization to work, we employ the \emph{parameter-shift} rule \cite{schuld2019evaluating, bergholm2018pennylane} to calculate the gradients of quantum functions. This method has been highly successful in VQC-based QML tasks \cite{bergholm2018pennylane, chen2020hybrid,chen2020qcnn,chen2020quantum,chen19,mari2019transfer,mitarai2018quantum}.
Given the knowledge of calculating the observable $\hat{P}$ of a quantum function,
\begin{equation}
f\left(x ; \theta_{i}\right)=\left\langle 0\left|U_{0}^{\dagger}(x) U_{i}^{\dagger}\left(\theta_{i}\right) \hat{P} U_{i}\left(\theta_{i}\right) U_{0}(x)\right| 0\right\rangle=\left\langle x\left|U_{i}^{\dagger}\left(\theta_{i}\right) \hat{P} U_{i}\left(\theta_{i}\right)\right| x\right\rangle,
\end{equation}
where $x$ is the classical input value (e.g. from input image array or post-measurement values of another quantum circuit) to the quantum circuit; $U_0(x)$ is the state preparation circuit to transform or encode the classical value $x$ into a quantum state; $i$ is the index of circuit parameter for which the gradient is to be evaluated;
and $U_i(\theta_i)$ represents the single-qubit rotation generated by the Pauli operators $X, Y$, and $Z$. It has been shown in the work \cite{mitarai2018quantum} that the gradient of this quantum function $f$ with respect to the parameter $\theta_i$ is
\begin{equation}
    \nabla_{\theta_i} f(x;\theta_i) = \frac{1}{2}\left[ f\left(x;\theta_i + \frac{\pi}{2}\right) - f\left(x;\theta_i - \frac{\pi}{2}\right)\right].
    \label{eq:quantum gradient}
\end{equation}
%
%
With the knowledge of calculating the quantum function gradients, it becomes straightforward to employ a variety of optimization algorithms developed by the classical ML community~\cite{ruder2016overview} and to train the whole hybrid architecture in an end-to-end fashion. In this work, the optimizer is chosen to be the RMSProp~\cite{Tieleman2012}, which is a variant of gradient-descent method with the feature of adaptive learning rate. The circuit parameters $\theta$ are updated according to:
\begin{subequations}
\begin{align}
        E\left[g^{2}\right]_{t} &= \alpha E\left[g^{2}\right]_{t-1}+ (1 - \alpha) g_{t}^{2}, \\ 
        \theta_{t+1} &= \theta_{t}-\frac{\eta}{\sqrt{E\left[g^{2}\right]_{t}}+\epsilon} g_{t},
\end{align}
\end{subequations} 
where $g_t$ is the gradient at step $t$ and $E\left[g^{2}\right]_{t}$ is the weighted moving average of the squared gradient with $E[g^2]_{t=0} = g_0^2$. The hyperparameters used in this work are: learning rate $\eta =0.01$, smoothing constant $\alpha = 0.99$, and $\epsilon = 10^{-8}$. 
\section{\label{sec:GraphConvAndVQC}Graph Convolution and VQC}
\begin{figure}[htbp]
\centering
\includegraphics[width=0.8\linewidth]{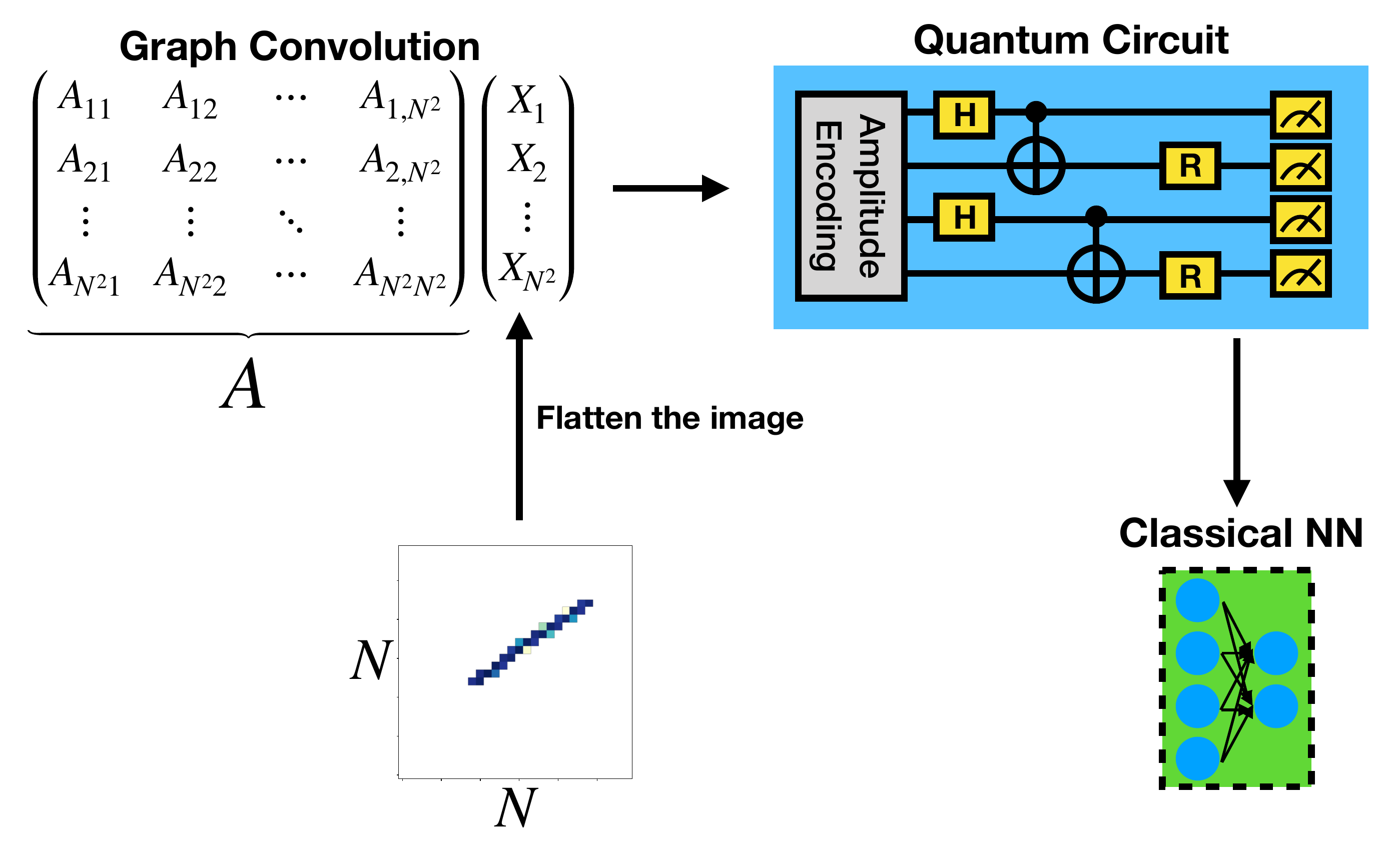}
\caption{{\bfseries Hybrid Quantum-Classical Graph Convolution.}
The proposed hybrid quantum-classical graph CNN contains three major components: 1) graph convolution 2) VQC, and 3) classical post-processing. The input image with dimension $N \times N$ will be operated first by the matrix $A$. Then, the processed input is encoded into a quantum state via amplitude encoding. Depending on the problem of interest, the quantum circuit portion may contain several different VQC blocks. The quantum measurement values from the final VQC block will be processed by a classical unit, which can be a neural network, to generate the logits of each class for classification.}
\label{quantumGCNDiagram}
\end{figure}

\begin{figure}[htbp]
\begin{center}
\begin{minipage}{10cm}
\Qcircuit @C=1em @R=1em {
\lstick{\ket{0}} & \multigate{9}{U(\mathbf{x})}   & \ctrl{1}   & \qw       & \qw      & \qw      & \qw      & \qw      & \qw      & \qw      & \qw     & \qw      & \targ    &\gate{R(\alpha_1, \beta_1, \gamma_1)} & \meter \qw \\
\lstick{\ket{0}} & \ghost{U(\mathbf{x})}         & \targ      & \ctrl{1}  & \qw      & \qw      & \qw      & \qw      & \qw      & \qw      & \qw      & \qw      & \qw      &\gate{R(\alpha_2, \beta_2, \gamma_2)} & \meter \qw \\
\lstick{\ket{0}} & \ghost{U(\mathbf{x})}         & \qw        & \targ     & \ctrl{1} & \qw      & \qw      & \qw      & \qw      & \qw      & \qw      & \qw      & \qw      &\gate{R(\alpha_3, \beta_3, \gamma_3)} & \meter \qw \\
\lstick{\ket{0}} & \ghost{U(\mathbf{x})}         & \qw        & \qw       & \targ    & \ctrl{1} & \qw      & \qw      & \qw      & \qw      & \qw      & \qw      & \qw      &\gate{R(\alpha_4, \beta_4, \gamma_4)} & \meter \qw \\
\lstick{\ket{0}} & \ghost{U(\mathbf{x})}         & \qw        & \qw       & \qw      & \targ    & \ctrl{1} & \qw      & \qw      & \qw      & \qw      & \qw      & \qw      &\gate{R(\alpha_5, \beta_5, \gamma_5)} & \meter \qw  \\
\lstick{\ket{0}} & \ghost{U(\mathbf{x})}         & \qw        & \qw       & \qw      & \qw      & \targ    & \ctrl{1} & \qw      & \qw      & \qw      & \qw      & \qw      &\gate{R(\alpha_6, \beta_6, \gamma_6)} & \meter \qw  \\
\lstick{\ket{0}} & \ghost{U(\mathbf{x})}         & \qw        & \qw       & \qw      & \qw      & \qw      & \targ    & \ctrl{1} & \qw      & \qw      & \qw      & \qw      &\gate{R(\alpha_7, \beta_7, \gamma_7)} & \meter \qw \\
\lstick{\ket{0}} & \ghost{U(\mathbf{x})}         & \qw        & \qw       & \qw      & \qw      & \qw      & \qw      & \targ    & \ctrl{1} & \qw      & \qw      & \qw      &\gate{R(\alpha_8, \beta_8, \gamma_8)} & \meter \qw \\
\lstick{\ket{0}} & \ghost{U(\mathbf{x})}         & \qw        & \qw       & \qw      & \qw      & \qw      & \qw      & \qw      & \targ    & \ctrl{1} & \qw      & \qw      &\gate{R(\alpha_9, \beta_9, \gamma_9)} & \meter \qw \\
\lstick{\ket{0}} & \ghost{U(\mathbf{x})}         & \qw        & \qw       & \qw      & \qw      & \qw      & \qw      & \qw      & \qw      & \targ    & \qw      & \ctrl{-9}&\gate{R(\alpha_{10}, \beta_{10}, \gamma_{10})} &  \meter \qw \gategroup{1}{3}{10}{14}{.7em}{--}
}
\end{minipage}
\end{center}
\caption[Variational quantum circuit architecture for the variational quantum classifier with amplitude encoding.]{{\bfseries Variational quantum circuit architecture for the classifier with amplitude encoding.} The first VQC block encodes the vector after the graph convolution operation. The $N$-dimensional vector is encoded via amplitude encoding into a $\log(N)$-qubit state. In this work, $N = 10$. The $U(\mathbf{x})$ is the quantum routine for amplitude encoding, which is described in \cite{mottonen2005transformation, Schuld2018InformationEncoding}. The parameters labeled with $\alpha_{i}$, $\beta_{i}$, and $\gamma_{i}$ are for optimization. The grouped box in the circuit may repeat several times to increase the number of parameters, subject to the capacity and capability of the available quantum devices or simulation software used for the experiments. The number of repeat is a hyperparameter and should be chosen carefully. In this work, the number of repeat is set to be $3$.
}
\label{Fig:VQCAmplitudeEncoding}
\end{figure}
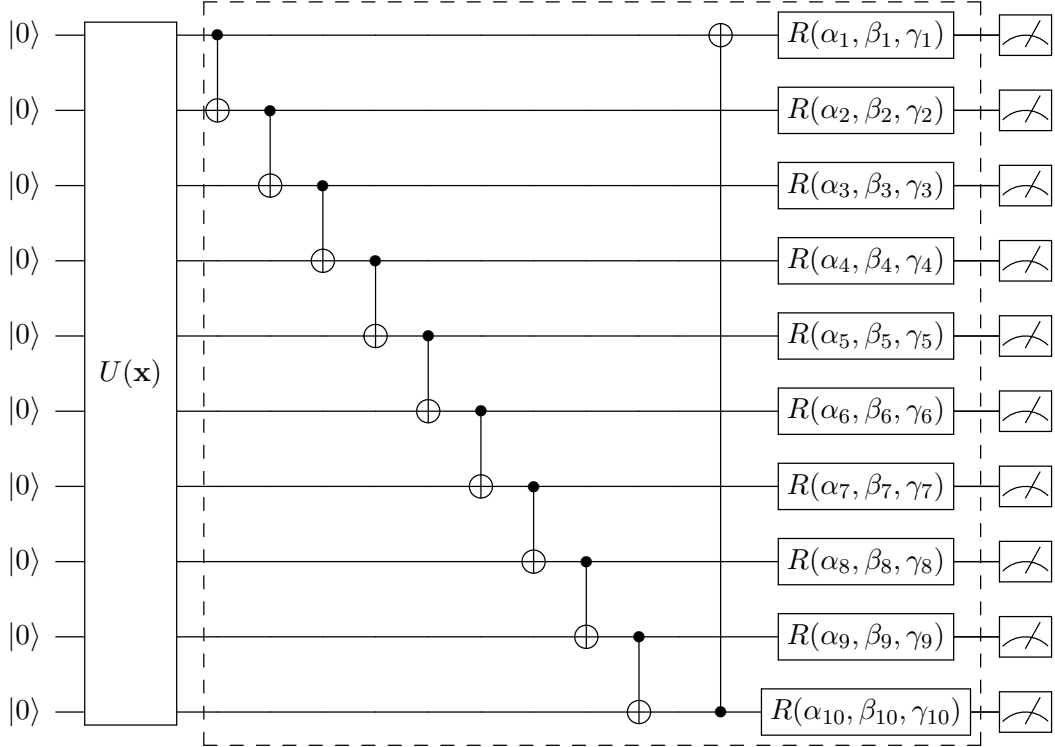

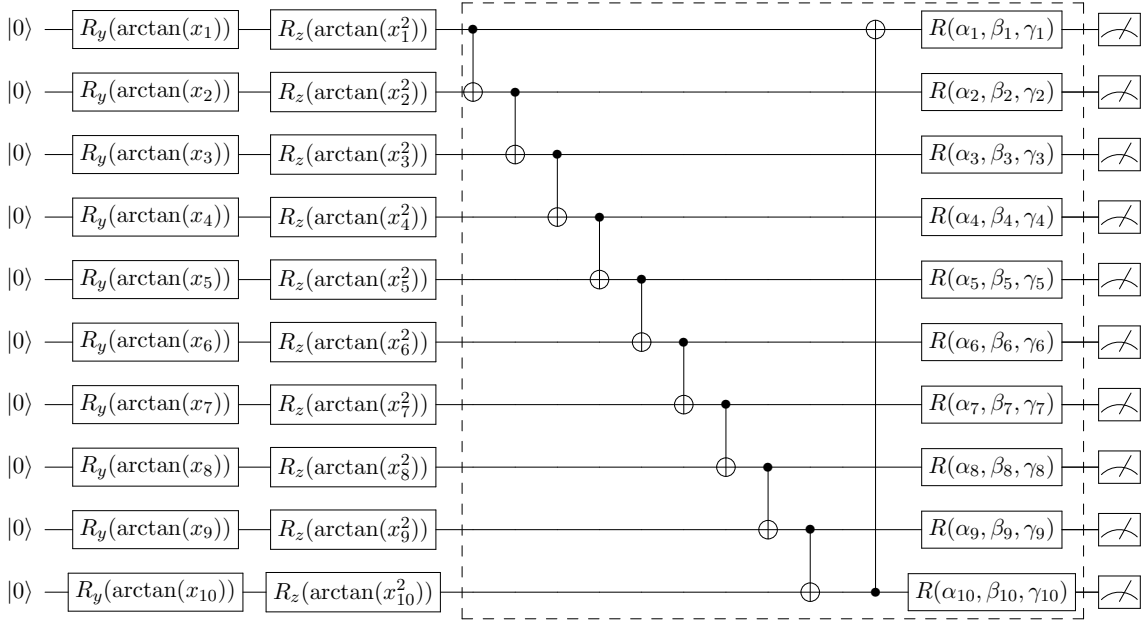
\begin{figure}[htbp]
\begin{center}
\scalebox{0.8}{

\begin{minipage}{10cm}
\Qcircuit @C=1em @R=1em {
\lstick{\ket{0}} & \gate{R_y(\arctan(x_1))} & \gate{R_z(\arctan(x_1^2))}         & \ctrl{1}   & \qw       & \qw      & \qw      & \qw      & \qw      & \qw      & \qw      & \qw     & \qw      & \targ    &\gate{R(\alpha_1, \beta_1, \gamma_1)} & \meter \qw \\
\lstick{\ket{0}} & \gate{R_y(\arctan(x_2))} & \gate{R_z(\arctan(x_2^2))}         & \targ      & \ctrl{1}  & \qw      & \qw      & \qw      & \qw      & \qw      & \qw      & \qw      & \qw      & \qw      &\gate{R(\alpha_2, \beta_2, \gamma_2)} & \meter \qw \\
\lstick{\ket{0}} & \gate{R_y(\arctan(x_3))} & \gate{R_z(\arctan(x_3^2))}         & \qw        & \targ     & \ctrl{1} & \qw      & \qw      & \qw      & \qw      & \qw      & \qw      & \qw      & \qw      &\gate{R(\alpha_3, \beta_3, \gamma_3)} & \meter \qw \\
\lstick{\ket{0}} & \gate{R_y(\arctan(x_4))} & \gate{R_z(\arctan(x_4^2))}         & \qw        & \qw       & \targ    & \ctrl{1} & \qw      & \qw      & \qw      & \qw      & \qw      & \qw      & \qw      &\gate{R(\alpha_4, \beta_4, \gamma_4)} & \meter \qw \\
\lstick{\ket{0}} & \gate{R_y(\arctan(x_5))} & \gate{R_z(\arctan(x_5^2))}         & \qw        & \qw       & \qw      & \targ    & \ctrl{1} & \qw      & \qw      & \qw      & \qw      & \qw      & \qw      &\gate{R(\alpha_5, \beta_5, \gamma_5)} & \meter \qw  \\
\lstick{\ket{0}} & \gate{R_y(\arctan(x_6))} & \gate{R_z(\arctan(x_6^2))}         & \qw        & \qw       & \qw      & \qw      & \targ    & \ctrl{1} & \qw      & \qw      & \qw      & \qw      & \qw      &\gate{R(\alpha_6, \beta_6, \gamma_6)} & \meter \qw  \\
\lstick{\ket{0}} & \gate{R_y(\arctan(x_7))} & \gate{R_z(\arctan(x_7^2))}         & \qw        & \qw       & \qw      & \qw      & \qw      & \targ    & \ctrl{1} & \qw      & \qw      & \qw      & \qw      &\gate{R(\alpha_7, \beta_7, \gamma_7)} & \meter \qw \\
\lstick{\ket{0}} & \gate{R_y(\arctan(x_8))} & \gate{R_z(\arctan(x_8^2))}         & \qw        & \qw       & \qw      & \qw      & \qw      & \qw      & \targ    & \ctrl{1} & \qw      & \qw      & \qw      &\gate{R(\alpha_8, \beta_8, \gamma_8)} & \meter \qw \\
\lstick{\ket{0}} & \gate{R_y(\arctan(x_9))} & \gate{R_z(\arctan(x_9^2))}         & \qw        & \qw       & \qw      & \qw      & \qw      & \qw      & \qw      & \targ    & \ctrl{1} & \qw      & \qw      &\gate{R(\alpha_9, \beta_9, \gamma_9)} & \meter \qw \\
\lstick{\ket{0}} & \gate{R_y(\arctan(x_{10}))} & \gate{R_z(\arctan(x_{10}^2))}         & \qw        & \qw       & \qw      & \qw      & \qw      & \qw      & \qw      & \qw      & \targ    & \qw      & \ctrl{-9}&\gate{R(\alpha_{10}, \beta_{10}, \gamma_{10})} &  \meter \qw \gategroup{1}{4}{10}{15}{.7em}{--}
}
\end{minipage}

}

\end{center}
\caption[Variational quantum circuit architecture for the variational quantum classifier.]{{\bfseries Variational quantum circuit block with variational encoding.} This encoding part includes $R_{y}$ and $R_{z} rotations$, parameterized by the rotation angles $\arctan(x_{i})$ and $\arctan(x_{i}^2)$ for each qubit. The parameters labeled with $\alpha_{i}$, $\beta_{i}$, and $\gamma_{i}$ are for optimization. The grouped box in the circuit may repeat several times to increase the number of parameters, subject to the capacity and capability of the available quantum devices or simulation software used for the experiments. The number of repeat is a hyperparameter and should be chosen carefully. In this work, the number of repeat is set to be $3$.
}
\label{Fig:VQCInterBlock}
\end{figure}

%
There are three major components in the proposed framework: 1) graph convolution, 2) VQC, and 3) classical post-processing. 
First, the original input image $X$ with the dimension $N \times N$ will be flattened into a one-dimensional vector, and the adjacency matrix $A$, which is defined in~\equationautorefname{\ref{eqn:adjacency_a_ij}}, will operate on it via matrix multiplication $n$ times, which is set to be $n = 2$ in this work. 
Then, the aggregated vector $A^{n}X$ will be encoded into a quantum state via amplitude encoding described in \sectionautorefname{\ref{sec:AmplitudeEncoding}} to maximally reduce the number of qubits used. 
The encoded quantum state then will go through several variational operations. Concretely, there are two VQCs separated by a tanh activation function. The first VQC (in \figureautorefname{\ref{Fig:VQCAmplitudeEncoding}}) is responsible for the amplitude encoding process and variational operations. The Pauli-$Z$ expectation values from the first VQC subsequently are encoded via the variational encoding method (described in \sectionautorefname{\ref{sec:VariationalQuantumCircuit}}) into the second VQC (see \figureautorefname{\ref{Fig:VQCInterBlock}}) and undergo variational operations.
Finally, the measured expectation values from the final VQC block (the second VQC in this study) are processed with a single-layer classical neural network to output the logits for each class.
For a schematic description of the framework, refer to \figureautorefname{\ref{quantumGCNDiagram}}.
%

\section{\label{sec:ExpAndResults}Experiments and Results}
This section presents the numerical simulation of the proposed QGCNN on the task of classifying different HEP events. The input data are in the dimension of $32 \times 32$. As noted, the dataset used in this study is the same as the one used in the previous work \cite{chen2020qcnn} (to aid comparison). The minor difference is that in this work, the image is padded into $32 \times 32$ for the amplitude encoding, which requires an input vector with a dimension of $2^n$.

Here, we compare our hybrid quantum-classical graph convolution models with three other related models (\tableautorefname{\ref{tab:comparison_num_parameters}}). We set the MLP as the baseline in this work. We also compare to results reported on the same dataset from previous work using QCNN and classical CNN with a similar number of parameters \cite{chen2020qcnn}.
\begin{table}[htbp]
\begin{threeparttable}
\centering
\begin{tabular}{|l|l|l|l|l|}
\hline
                     & QGCNN & MLP & QCNN \tnote{a}\quad \quad & CNN \tnote{a} \  \\ \hline
Number of Parameters & $202$   & $131458$  & $472$   & $498$  \\ \hline
\end{tabular}
\begin{tablenotes}
            \item[a] QCNN and CNN are from the work \cite{chen2020qcnn}.
\end{tablenotes}
\end{threeparttable}
\caption{{\bfseries Comparison of the number of parameters in different models.} The proposed QGCNN is compared with other related architectures in terms of the number of parameters. The MLP is the baseline used in this work with a single hidden layer. The QCNN and CNN are from the previous work \cite{chen2020qcnn}. 
}
\label{tab:comparison_num_parameters}
\end{table}
%

%
For the proposed QGCNN, we first process the input image $X$ with the adjacency matrix $A$. The transformed input $A^{2}X$  then is encoded into the quantum circuit via amplitude encoding. There are two VQC blocks in the model, separated by a quantum measurement layer. The first quantum circuit block (shown in \figureautorefname{\ref{Fig:VQCAmplitudeEncoding}}) is in conjunction with the amplitude encoding routine and has $10 \times 3 \times 3 = 90$ parameters. The second quantum block (shown in \figureautorefname{\ref{Fig:VQCInterBlock}}) encodes the measured values from the first block then operates on another $10 \times 3 \times 3 = 90$ parameters. The measured expectation values from the second block are processed with a single-layer classical neural network (with $10 \times 2 + 2 = 22$ parameters) to output two-dimensional logits for binary classification. Therefore, the total number of parameters in the QGCNN model is $90 + 90 + 22 = 202$.
The MLP baseline used in this work is: $1024 \times 128 + 128 + 128 \times 2 + 2 = 131458$. 
%
%
The software used for this work includes PyTorch \cite{paszke2019pytorch}, PennyLane \cite{bergholm2018pennylane}, and Qulacs \cite{suzuki2020qulacs}.
\begin{figure}[htbp]
\centering
\includegraphics[width=1.\linewidth]{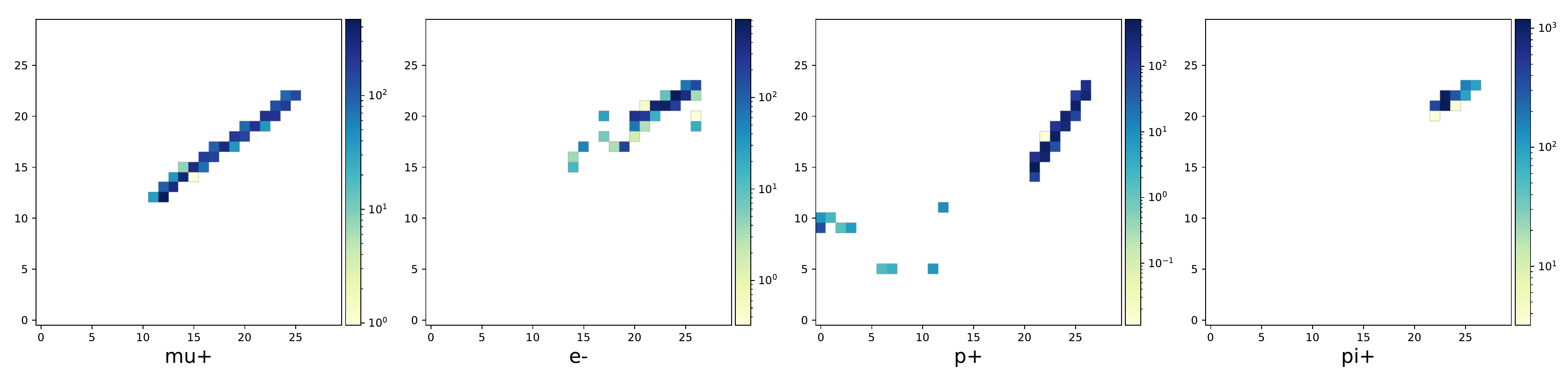}
\caption{Examples of scaled images of simulated particle activities ($\mu^+$, $\pi^+$, $p$, $e^-$) in a LArTPC detector. These are the images used in the QGCNN experiments. In the experiment, we pad $0$ to the image, so the dimension of these images is $32 \times 32$, which is for amplitude encoding with a $10$-qubit quantum circuit.}
\label{scaled_examples}
\end{figure}
\subsection{Muon versus Electron}
\begin{figure}[htbp]
\centering
\includegraphics[width=1\linewidth]{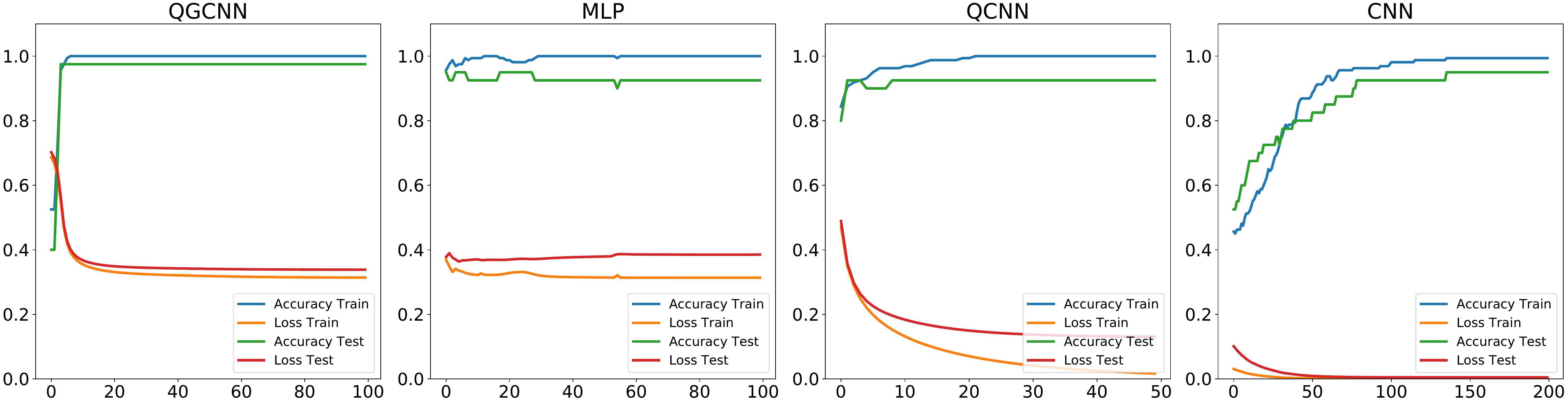}
\caption{{\bfseries Result: Hybrid Quantum-Classical GCN on binary classification of muon versus electron.} Comparison of performance between different architectures in the task of binary classification of muon versus electron. For a better comparison, results from the previous work on QCNN versus CNN are included \cite{chen2020qcnn}. In \tableautorefname{\ref{tab:comparison_num_parameters}}, we present the number of parameters in differing architectures. 
}
\label{comparison_results_mu_electron}
\end{figure}

\figureautorefname{\ref{comparison_results_mu_electron}} and \tableautorefname{\ref{tab:results_comparison_mu_e}} show the results of the classification between $\mu^+$ and $e^-$. A $\mu^+$ is a track-like particle, while an $e^-$ produces electromagnetic showers that are spatially extended. The patterns from these two particles are rather distinctive visually. For better comparison, we include the QCNN and CNN from the previous work \cite{chen2020qcnn}. In this experiment, we can observe comparable performance in the four different architectures in terms of testing accuracies. While QGCNN has a slightly better testing accuracy, the margin is not significant because this particular task is not too difficult. If we consider the number of parameters, it is evident that QGCNN performs better as it requires fewer model parameters. 
%
%
\begin{table}[htbp]
\centering
\begin{threeparttable}
\begin{tabular}{|l|l|l|l|l|}
\hline
     & Training Accuracy & Testing Accuracy & Training Loss & Testing Loss \\ \hline
QGCNN & $100\%$              & $97.5\%$             & $0.3138$            & $0.3384$           \\ \hline
MLP  & $100\%$           & $92.5\%$         & $0.3134$      & $0.3852$           \\ \hline
QCNN \tnote{a} \  & $100\%$              & $92.5\%$             & $0.017$            & $0.13$           \\ \hline
CNN \tnote{a} & $99.38\%$              & $95\%$             & $0.0002$            & $0.0046$           \\ \hline
\end{tabular}
\begin{tablenotes}
            \item[a] QCNN and CNN are from the work \cite{chen2020qcnn}.
\end{tablenotes}
\end{threeparttable}
\caption{Performance comparison between QGCNN and other QML architectures on the binary classification between $\mu^+$ versus $e^-$.}
\label{tab:results_comparison_mu_e}
\end{table}
\subsection{Muon versus Proton}
\begin{figure}[htbp]
\centering
\includegraphics[width=1\linewidth]{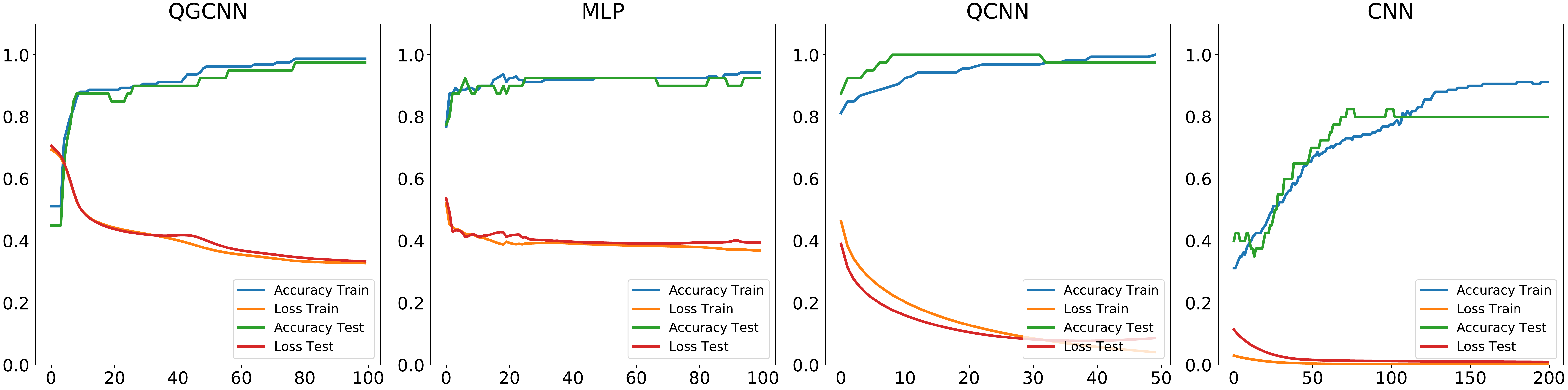}
\caption{{\bfseries Result: Hybrid Quantum-Classical GCN on binary classification of muon versus proton.} Comparison of performance between different architectures in the task of muon versus proton binary classification. Results from the previous work with QCNN versus CNN are included for better comparison \cite{chen2020qcnn}. \tableautorefname{\ref{tab:comparison_num_parameters}} also presents the number of parameters in differing architectures.
}
\label{comparison_results_mu_proton}
\end{figure}

\figureautorefname{\ref{comparison_results_mu_proton}} and \tableautorefname{\ref{tab:results_comparison_mu_p}} show the results of the classification between $\mu^+$ and $p$. 
Because a proton's mass is much heavier than a muon, it has higher energy deposition per unit length and encounters less multiple Coulomb scattering when it passes the detector. As a result, a proton's track has higher pixel intensity and is straighter than that of a muon.
Here, we include the QCNN and CNN from the previous work \cite{chen2020qcnn} for better comparison. In this experiment, QGCNN presents a comparable performance to QCNN in terms of testing accuracies. However, QGCNN requires fewer parameters than QCNN. We also observe that QGCNN has significantly better performance than the MLP baseline and CNN. Specifically, the number of parameters in QGCNN is much lower than that of the MLP.
%
%
\begin{table}[htbp]
\centering
\begin{threeparttable}
\begin{tabular}{|l|l|l|l|l|}
\hline
     & Training Accuracy & Testing Accuracy & Training Loss & Testing Loss \\ \hline
QGCNN & $98.75\%$              & $97.5\%$             & $0.3283$            & $0.3344$           \\ \hline
MLP  & $94.38\%$         & $92.5\%$             & $0.3688$            & $0.3950$           \\ \hline
QCNN \tnote{a} \ & $100.00\%$        & $97.5\%$             & $0.041$       & $0.087$           \\ \hline
CNN  \tnote{a} & $91.25\%$         & $80\%$             & $0.002$            & $0.01$           \\ \hline
\end{tabular}
\begin{tablenotes}
            \item[a] QCNN and CNN are from the work \cite{chen2020qcnn}.
\end{tablenotes}
\end{threeparttable}
\caption{Performance comparison between QGCNN and other QML architectures on the binary classification between $\mu^+$ versus $p$.}
\label{tab:results_comparison_mu_p}
\end{table}

\subsection{Muon versus Charged Pion}
\begin{figure}[htbp]
\centering
\includegraphics[width=1\linewidth]{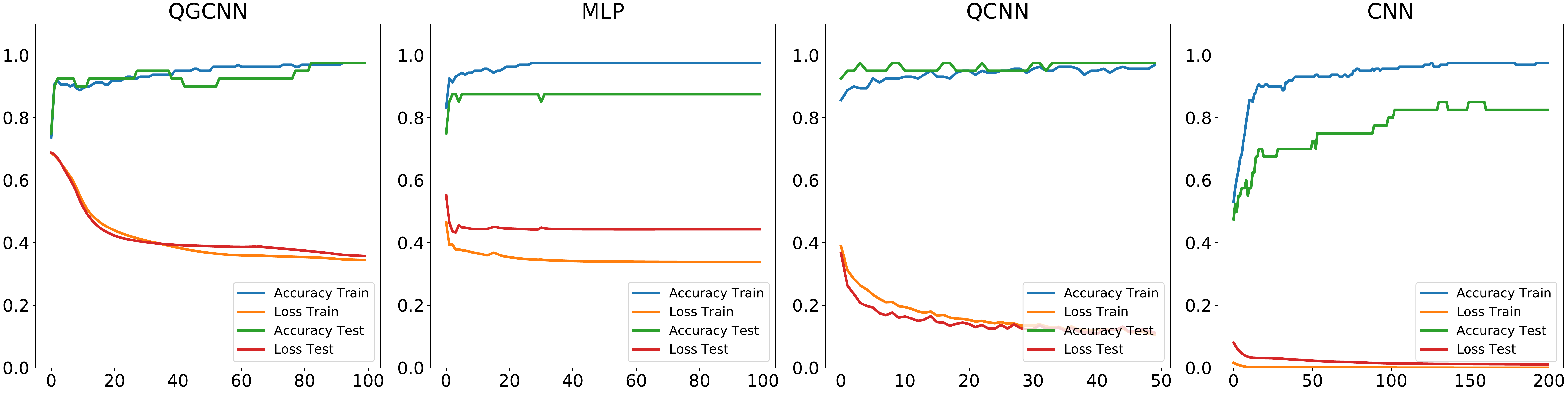}
\caption{{\bfseries Result: Hybrid Quantum-Classical GCN on binary classification of muon versus charged pion.} Comparison of performance between different architectures in the task of muon versus charged pion binary classification. Results from the previous work on QCNN versus CNN are included for better comparison \cite{chen2020qcnn}. \tableautorefname{\ref{tab:comparison_num_parameters}} also presents the number of parameters in differing architectures.
}
\label{comparison_results_mu_charged_pion}
\end{figure}

\figureautorefname{\ref{comparison_results_mu_charged_pion}} and \tableautorefname{\ref{tab:results_comparison_mu_pi}} depict the results of the classification between $\mu^+$ and $\pi^+$. A charged pion behaves similarly to a muon because their masses are closer. The main difference is that the $\pi^+$ experiences additional nuclear interactions during its passage in the detector, often leading to a large-angle ``kink'' along its main trajectory. For better comparison, we include the QCNN and CNN from the previous work \cite{chen2020qcnn}. In this experiment, QGCNN illustrates a comparable performance to QCNN in terms of testing accuracies. However, QGCNN requires fewer parameters than QCNN. We also observe that QGCNN offers significantly better performance than the MLP baseline and CNN. In particular, the number of parameters in QGCNN is much lower than that of MLP.
%
%
\begin{table}[htbp]
\centering
\begin{threeparttable}
\begin{tabular}{|l|l|l|l|l|}
\hline
     & Training Accuracy & Testing Accuracy & Training Loss & Testing Loss \\ \hline
QGCNN & $97.5\%$              & $97.5\%$             & $0.3448$            & $0.3574$           \\ \hline
MLP  & $97.5\%$          & $87.5\%$         & $0.3384$      & $0.4431$           \\ \hline
QCNN \tnote{a} \ & $96.88\%$         & $97.5\%$         & $0.1066$      & $0.1121$     \\ \hline
CNN  \tnote{a} & $97.5\%$              & $82.5\%$             & $0.0006$            & $0.0116$           \\ \hline
\end{tabular}
\begin{tablenotes}
            \item[a] QCNN and CNN are from the work \cite{chen2020qcnn}.
\end{tablenotes}
\end{threeparttable}
\caption{Performance comparison between QGCNN and other QML architectures on the binary classification between $\mu^+$ versus $\pi^+$.}
\label{tab:results_comparison_mu_pi}
\end{table}

\section{\label{sec:Discussion}Discussion}
\subsection{Quantum Graph Encoding}
In this work, we combine the classical aggregation matrix and quantum amplitude encoding to perform the classification. It is interesting to investigate the possibility of performing the graph convolution step with the quantum operation. For example, the work in \cite{thabet2020laplacian} introduced a method of encoding the Laplacian eigenmap with VQC.
In the work of \cite{ma2019variational}, the authors propose a graph embedding method based on variational circuits to deal with knowledge graphs.
\subsection{Future Applications}
%
Graph CNNs have been studied extensively among the classical ML community \cite{kipf2016semi, henaff2015deep, wu2020comprehensive, zhou2018graph, zhang2018graph, sun2020graph, liu2020efficient}. Several important applications have been demonstrated, for example, social network prediction \cite{kipf2016semi, hamilton2017inductive}, traffic problems \cite{rahimi2018semi, cui2019traffic}, recommender systems \cite{ying2018graph, berg2017graph}, graph representation \cite{ying2018hierarchical}, graph generation \cite{bojchevski2018netgan, you2018graph}, computational chemistry \cite{duvenaud2015convolutional}, drug development \cite{sun2020graph}, and modeling physical dynamics \cite{battaglia2016interaction} (to name a few). Our hybrid quantum-classical model is expected to be applicable to most of these aforementioned areas. Another interesting will be to investigate the potential quantum advantages of using different quantum architectures in diverse real-world scenarios. 
\subsection{Quantum Machine Learning in High Energy Physics}
%
%
Applying QML in HEP data analysis is an emerging field, yet there are several related works in this area. These works \cite{wu2020application, terashi2020event, trenti2020quantum} focus on event classification with QML models. For example, \cite{wu2020application, terashi2020event} used similar VQC models to study the classification problems in HEP. However, the input dimension is limited. In \cite{trenti2020quantum}, the authors used a quantum-inspired classical algorithms, called ``tree tensor network,'' to study the classification problem. The tensor network formulation has a direct corollary in the quantum circuit model. The underlying relationships of these models and other purely VQC-based models deserve further investigation. 
Recent advances in building more sophisticated QML models also have led to the successful demonstration of applying QCNN to HEP event classification \cite{chen2020qcnn}.
Meanwhile, using classical graph neural networks in HEP research has become popular \cite{shlomi2020graph}. The work in \cite{tuysuz2020quantum} proposed a quantum graph model for particle track reconstruction. However, our approach differs from this one. We employ a classical graph convolutional operation on a regular graph that is an image then encode the transformed vector into a quantum state via amplitude encoding to reduce the number of qubits used.
%
For more in-depth discussions regarding QML in HEP, refer to recent reviews \cite{guan2020quantum, sharma2020quantum}.
\section{\label{sec:Conclusion}Conclusion}
This work demonstrates a hybrid quantum classical QCNN that extends the power of graph convolution operation to enhance the features gleaned from input data and the capability of quantum superposition to greatly reduce the number of model parameters used. Notably, we numerically show the significantly superior performance in terms of testing accuracies over using classical MLP, classical CNN, and recent QCNN methods. These results indicate the potential benefits such capabilities can bring to the HEP community and other scientific areas in the quantum era.

\begin{acknowledgments}
This work is supported by the U.S.\ Department of Energy, Office of Science, Office of High Energy Physics program under Award Number DE-SC-0012704 and Brookhaven National Laboratory's Laboratory Directed Research and Development Program \#20-024.
\end{acknowledgments}

\appendix


\bibliographystyle{ieeetr}
\bibliography{apssamp,bib/gcn,bib/tools,bib/vqc,bib/qml_examples,bib/quantum_graph,bib/nisq,bib/qc,bib/qml_hep_related,bib/ml_hep}

\end{document}